\documentclass[letterpaper]{article} 
\usepackage{aaai24}  
\usepackage{times}  
\usepackage{helvet}  
\usepackage{courier}  
\usepackage[hyphens]{url}  
\usepackage{graphicx} 
\usepackage{natbib}  
\usepackage{caption} 
\usepackage{amssymb}
\usepackage{amsmath}

\usepackage{algorithm}
\usepackage{algorithmic}

\usepackage{newfloat}
\usepackage{listings}
\DeclareCaptionStyle{ruled}{labelfont=normalfont,labelsep=colon,strut=off} 
\floatstyle{ruled}
\newfloat{listing}{tb}{lst}{}
\floatname{listing}{Listing}
\usepackage{multirow}  
\usepackage{booktabs} 

\title{Towards Continual Learning Desiderata via HSIC-Bottleneck Orthogonalization and Equiangular Embedding}
\author{
    Depeng Li\textsuperscript{\rm 1}\thanks{These authors contributed equally.},
    Tianqi Wang\textsuperscript{\rm 1}\footnotemark[1],
    Junwei Chen\textsuperscript{\rm 1},
    Qining Ren\textsuperscript{\rm 1},
    Kenji Kawaguchi\textsuperscript{\rm 2},
    Zhigang Zeng\textsuperscript{\rm 1}\thanks{Corresponding author.}
}
\affiliations{
    \textsuperscript{\rm 1}School of Artificial Intelligence and Automation, Huazhong University of Science and Technology\\
    \textsuperscript{\rm 2}School of Computing, National University of Singapore\\

    \{dpli, tianqiwang, junwei\_chen, qiningren, zgzeng\}@hust.edu.cn,  kawaguch@csail.mit.edu
}

\usepackage{bibentry}

\begin{document}

\maketitle

\begin{abstract}
Deep neural networks are susceptible to catastrophic forgetting when trained on sequential tasks. Various continual learning (CL) methods often rely on exemplar buffers or/and network expansion for balancing model stability and plasticity, which, however, compromises their practical value due to privacy and memory concerns. Instead, this paper considers a strict yet realistic setting, where the training data from previous tasks is unavailable and the model size remains relatively constant during sequential training. To achieve such desiderata, we propose a conceptually simple yet effective method that attributes forgetting to layer-wise parameter overwriting and the resulting decision boundary distortion. This is achieved by the synergy between two key components: HSIC-Bottleneck Orthogonalization (HBO) implements non-overwritten parameter updates mediated by Hilbert-Schmidt independence criterion in an orthogonal space and EquiAngular Embedding (EAE) enhances decision boundary adaptation between old and new tasks with predefined basis vectors. Extensive experiments demonstrate that our method achieves competitive accuracy performance, even with absolute superiority of zero exemplar buffer and 1.02$\times$ the base model. 
\end{abstract}

\section{Introduction}
Current deep learning models have shown promising performance in various fields, but they lack the ability of continual learning (CL) that humans possess~\cite{kang2022forget, smith2023coda}. CL entails progressively acquiring knowledge from sequentially presented tasks, with access to only current task data and no past data~\cite{li2023CRNet}. As a result, directly retraining a well-trained model on new task data using stochastic gradient descent (SGD) leads to the well-known phenomenon of \textit{catastrophic forgetting}~\cite{mccloskey1989catastrophic}, which refers to abrupt and significant performance degradation on previously learned tasks.

\begin{figure}[t]
	\centering
	\includegraphics[width=1.0\columnwidth]{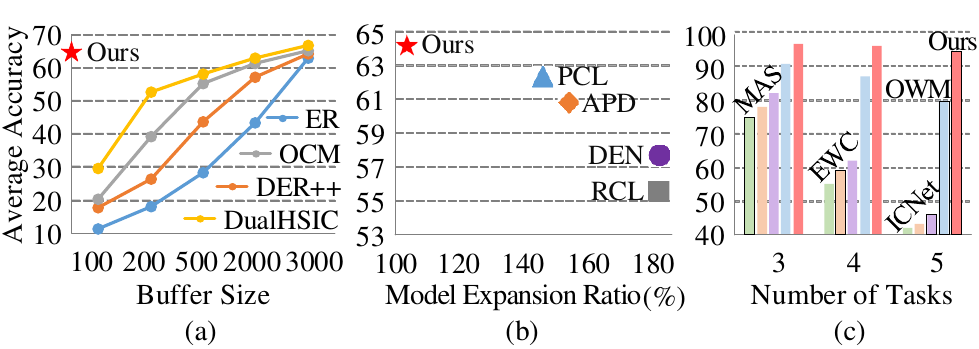} 
	\caption{Comparison between our method and representative CL approaches. (a) Rehearsal-based ones are often sensitive to buffer sizes. (b) Some architecture-based ones scale rapidly during sequential training. (c) Most regularization-based ones struggle with the stability-plasticity dilemma whose performance is not satisfactory in the class-IL (hybrid with (a) or/and (b) excluded). By contrast, our method reaches multiple CL desiderata simultaneously.}
	\label{Fig_Intro}
\end{figure}

Recent works have experienced a remarkable surge in addressing catastrophic forgetting~\cite{wang2021training, tong2022incremental, zhou2022model}. However, it is noteworthy that the merits of CL come with costs.
\textit{Rehearsal-based approaches}, as the mainstay of CL, explicitly buffer a small subset of past samples and retrain them with those from a new task jointly~\cite{liu2020mnemonics, hayes2020remind, bonicelli2022effectiveness, guo2022online, luo2023class}. Critically, these methods pose a threat to data privacy and often decline performance as buffer size decreases, as depicted in Figure 1(a). \textit{Architecture-based approaches} dynamically modify the network architecture to accommodate knowledge needed for new tasks~\cite{serra2018overcoming, ke2021achieving, yang2022dynamic, hu2023dense}. In particular, network expansion involves adding a sub-network for each task and utilizing aggregated feature representation for final prediction~\cite{yan2021dynamically, wang2022foster}. As shown in Figure 1(b), their model size expands rapidly as the number of tasks grows, which should be counted into the memory budget for a fair comparison~\cite{zhou2022model}. \textit{Regularization-based approaches} penalize parameter variations over an over-parameterized network, where each network parameter is associated with weight importance \cite{PNAS2017EWC, NMI2019OWM, wolczyk2022continual}. However, the performance of these methods that do not store any past data is yet unsatisfactory, especially in the class-incremental learning (class-IL) scenario, which addresses the most common problem of incrementally learning new classes without the provision of test-time task identities ~\cite{zhuang2022acil, wang2022dualprompt} (see Figure 1(c)). 

In summary, for alleviating catastrophic forgetting, many CL methods prioritize accuracy performance to the detriment of other fronts. This motivates us to find new methods against forgetting while satisfying multiple CL desiderata: 
(i)~\textit{It should no longer access the training data of previous tasks}. While keeping prior observations demonstrates superior ability in combating forgetting, reliance on rehearsal buffers may not be memory-efficient \cite{WACV2020DMC, li2023IF2Net, luo2023class}. Importantly, this involves violating practical constraints such as privacy and security issues~\cite{shokri2015privacy}, which are common in domains like federated learning~\cite{qi2023better}. 
(ii)~\textit{It should remain the model size relatively unchanged during sequential training.} Instead of buffering data, storing backbones from the history (e.g., network expansion) pushes the performance towards the upper bound progressively~\cite{yan2021dynamically, zhou2022model, hu2023dense}. The drawback lies in the growth can be computationally expensive and it is intractable to customize the growth quota compactly matching the difficulty of a newly arriving task~\cite{dai2019stochastic}. 
(iii)~\textit{It should strike a balance between the stability and plasticity}~\cite{NIPS2019RPS-Net, kim2023stability}, ensuring not only the persistent knowledge retention of past tasks but also the sufficient capacity to accommodate new ones. Intuitively, it is difficult, if not impossible, for pure parameter regularization to achieve such balance via the current learning paradigm. 

With these considerations, a straightforward attempt is to seek alternative solutions to parameter regularization, with extra care to not infringe its inherent merits. Inspired by this insight, in this paper, we develop a drastically different training objective that \textit{recasts} a representative regularizer in reaching multiple CL desiderata. Termed \textit{CLDNet}, we avoid the conventional cross-entropy loss and instead incorporate statistical dependency and distance metric, achieving a better stability-plasticity trade-off in a rehearsal-free and minimal-expansion fashion.
To this end, we decompose the CL process into dual problem-solving: (1) How to address the layer-wise parameter overwriting due to the scarcity of prior task data? (2) How to mitigate the inter-/intra-task confusion caused by distortions in decision boundaries?  

To approach the first question, we leverage the interplay of Hilbert-Schmidt independence criterion (HSIC) and orthogonal projection, hence the name HSIC-Bottleneck Orthogonalization (HBO). Taking a close look at both: HSIC is a non-parametric kernel-based technique utilized to assess the statistical (in)dependence of different layers, which has been widely adopted for various learning tasks~\cite{wang2021learning} but is under-investigated in CL community~\cite{wang2023dualhsic}; And a basic idea behind the orthogonal projection is to regularize gradient update directions that do not disturb the weights of previous tasks~\cite{NMI2019OWM}. 
Based on them, the introduced HBO implements non-overwritten parameter updates facilitated by the HSIC-bottleneck training in an orthogonal space, where one can exploit readily available gradient updates by measuring nonlinear dependencies between the inputs and outputs. It requires no access to or storing of previous data, no architecture growth, and no awareness of test-time task identities.

To address the aforementioned second question, we draw inspiration from the recently proposed equiangular basis vectors (EBVs)~\cite{shen2023equiangular}. Unlike the trainable fully-connected layer with softmax, the EBVs is parameter-free since its learning objective is to minimize the spherical distance of learned representations with predefined basis vectors. This ensures that the trainable parameters of deep neural networks are constant even with the growth of tasks. Though attractive, it remains unclear how to extend EBVs to CL. To bridge this gap, we design an EquiAngular Embedding (EAE) component that sits atop HBO. During each CL session, we vectorize the embedding of HBO output and optimize it towards its class-specific equiangular basis vector in a scalable manner. Compared to the standard classification, EAE exhibits a stronger discriminative ability through the tailored distance metric, thereby enhancing decision boundary adaptation between old and new tasks. 

Benefiting from the synergy between HBO and EAE, our CLDNet reaches multiple CL desiderata. Empirical evaluation across a range of widely used benchmark datasets  demonstrates the superiority of our approach in terms of exemplar buffers, network expansion, and competitive performance. On CIFAR-100, for instance, CLDNet outperforms the state-of-the-art rehearsal-based baseline by 7.54\% with a minimal expansion ratio of 1.02$\times$ and buffer size of 0.



\section{Related Work}

Prior works to address catastrophic forgetting in CL can be broadly divided into three categories, from which we discuss a selection of representatives and focus on ones mostly related to our work. Rehearsal-based approaches preserve model stability by keeping a memory buffer of past samples at either input layer~\cite{liu2020mnemonics, bonicelli2022effectiveness} or hidden layer~\cite{yang2023neural} for joint training. GDumb~\cite{prabhu2020gdumb} employs a limited memory to buffer data in the order of arrival and dynamically replaces previously stored data. Rather than pixel-level exemplars, i-CTRL~\cite{tong2022incremental} is founded on compact and structured representations, while REMIND~\cite{hayes2020remind} stores hidden representations and reconstructs synthesized features for rehearsal. 

Architecture-based approaches either isolate model parameters~\cite{serra2018overcoming, ke2021achieving} or expand additional network branches~\cite{yang2022dynamic, hu2023dense}. PNN~\cite{rusu2016progressive} gradually adds new branches for all layers horizontally. 
RPS-Net~\cite{NIPS2019RPS-Net} uses parallel modules at each layer where a possible searching space is formed to contain previous task-specific knowledge. Methods such as PCL~\cite{AAAI2021PCL}, DER~\cite{yan2021dynamically}, and FOSTER~\cite{wang2022foster} acquire sufficient model plasticity by allocating a sub-network per task.

Regularization-based approaches mainly employ penalty terms to impose constraints on weights deemed important for old tasks~\cite{WACV2020DMC, li2023CLSNet}. The pioneering work was conducted by EWC~\cite{PNAS2017EWC}, followed by SI \cite{ICML2017SI}, and MAS \cite{ECCV2018MAS}. 
On the other hand, orthogonal projection-driven methods address forgetting by designing network parameter updating rules~\cite{farajtabar2020orthogonal, li2023MVCNet}. 
OWM~\cite{NMI2019OWM} constructs an orthogonal projector such that its gradient updates only occur in directions orthogonal to the input of previous tasks. In the multi-head setting (e.g., a separate classifier per task), GPM~\cite{saha2021gradient} stores the bases of core gradient space while FS-DGPM~\cite{deng2021flattening} further predicts the importance of such bases aided with a rehearsal buffer. Our work is also built on the basis of orthogonal projection but is very different from existing approaches as its training objective consists of the statistical dependency and distance metric, achieving a better stability-plasticity trade-off in a rehearsal-free and minimal-expansion fashion.  

Among the latest CL approaches, our work is closely related to AOP~\cite{guo2022adaptive}, OCM~\cite{guo2022online}, and DualHSIC~\cite{wang2023dualhsic}. (1) AOP aims to improve OWM itself by introducing a rule of expectation serving to strengthen orthogonal projectors.
By contrast, inspired by HSIC~\cite{ma2020hsic}, we reformulate the OWM process as dependence minimization or maximization problems in a unified way. To the best of our knowledge, it is the first orthogonal projector that utilizes HSIC for CL. (2) OCM turns to a complicated contrastive learning proxy over \textit{two models} to maximize the mutual information (MI), while CLDNet's HBO detects nonlinear dependencies with the advantage of easy empirical estimation over MI. (3) DualHSIC realizes CL by considering the inter-task relationship into task-specific and task-invariant knowledge, while CLDNet directly addresses forgetting via non-parameter overwriting and decision boundary adaptation. Note that both OCM and DualHSIC rely on rehearsal buffers that we do not. Interestingly, we observe that both OCM and DualHSIC require the additional trainable projection head; Instead, we bring a parameter-free classifier to CL, which enhances model decision ability as evidenced by the recent study~\cite{shen2023equiangular}. Therefore, our work differs significantly in terms of motivation and methodology.

\section{Preliminaries}


\paragraph{Hilbert-Schmidt Independence Criterion}
HSIC is a kernel-based measure of dependence between random variables~\cite{gretton2005measuring}. With it, one can transform many existing learning tasks into statistical independence minimization (or maximization) problems, much akin to MI. Unlike the indirect variational bounds on MI~\cite{poole2019variational}, it can be directly estimated given a finite number of observations. Therefore, it has
been used in a variety of applications for machine learning~\cite{ma2020hsic, wang2021revisiting, li2021self, kawaguchi2023does}.

Formally, given two random variables $X$ and $Y$ jointly drawn from probability distribution $P_{XY}$, HSIC identifies their dependency by first taking a nonlinear feature transformation of each, say $\phi: X \rightarrow \mathcal{H}$ and $\psi: Y \rightarrow \mathcal{G}$, with $h$ and $g$ being kernel functions in the $\mathcal{H}$ and $\mathcal{G}$ Hilbert spaces respectively. Let $(X^{'}, Y^{'})$ be independent copies of $(X, Y)$:
\begin{equation}\label{HSIC_E}
	\begin{split}
		{\rm HSIC}(P_{XY},&\mathcal{H},\mathcal{G}) = \mathbb{E}_{XYX^{'}Y^{'}}[h(X,X^{'})g(Y,Y^{'})] \\
		& + \mathbb{E}_{XX^{'}}[h(X,X^{'})]\mathbb{E}_{YY^{'}}[g(Y,Y^{'})]
		\\
		& - 2\mathbb{E}_{XY}[\mathbb{E}_{X^{'}}[h(X,X^{'})]\mathbb{E}_{Y^{'}}[g(Y,Y^{'})]] 
	\end{split}
\end{equation}

\noindent The formulation suggests that HSIC captures nonlinear dependencies between $X$ and $Y$, with the magnitude of the index indicating the strength of associations.

To render HSIC a practical measure for learning tasks, it has proven to be easily evaluated from mini-batches of data~\cite{gretton2005measuring, song2012feature}. Given $n$ i.i.d. samples $\mathcal{D} = \{(x_i,y_i)\}_{i=1}^n$ drawn from $P_{XY}$, the empirical estimation of HSIC is:
\begin{equation}\label{HSIC_e}
	\widehat{{\rm HSIC}}(\mathcal{D},\mathcal{H},\mathcal{G}) = (n-1)^{-2}{\rm tr}(HKGK)
\end{equation}

\noindent where ${\rm tr}(\cdot)$ is the trace operator, $H_{ij} = h(x_i,x_j)$ and $G_{ij} = g(y_i,y_j)$ are kernel matrices,
and $K = I_n - \frac{1}{n}1_{n}1_{n}^\mathrm{T}$ is the centering matrix. In our CLDNet, we evaluate the HSIC term in this empirical expression and denote it as ${\rm HSIC}(X, Y)$.

\paragraph{Equiangular Basis Vectors}
Although a trainable classifier with softmax cross-entropy remains the predominant approach for classification tasks, the potential of using a preassigned classifier has been explored~\cite{mettes2019hyperspherical, zhou2022forward}. The general idea is to replace the classifier weight with the class prototype, e.g., fixing the classifier as a Hadamard matrix~\cite{hoffer2018fix}, regular polytope~\cite{pernici2021class}, and simplex ETF~\cite{yang2022inducing}. In this paper, we are encouraged by the recently proposed equiangular basis vectors (EBVs)~\cite{shen2023equiangular}, characterized by simplicity and effortless implementation over prior studies.

Prior to learning, the EBVs assigns $C$ $d$-dimensional basis vectors on the surface of a unit hypersphere $S^d \in \mathbb{R}^d$, denoted by the set $\mathcal{W} = \{w_c\}_{c=1}^C$. In the predefined process, these basis vectors are pairwise
separated by the common angle $\gamma \in [0,1)$, which satisfies:
\begin{equation}\label{EBV_Def}
	-\gamma \leq \frac{w_i^\mathrm{T}\cdot w_j}{\Vert w_i\Vert\Vert w_j\Vert} \leq \gamma, \forall w_i,w_j\in \mathcal{W}, i\neq j
\end{equation}

\noindent where $\Vert \cdot \Vert$ is the Euclidean norm. Let $\varphi$ denote the spherical distance function. The EBVs produces a distribution over classes based on softmax: 
\begin{equation}\label{EBV_Dis}
	p(y=y_c|z) = \frac{{\rm exp}(\varphi(z,w_c))}{\sum_{c'=1}^C{\rm exp}(\varphi(z,w_{c'}))}
\end{equation}

\noindent where $z\in\mathbb R^{d}$ denotes the learned feature representation by a backbone network given the input $x$ and $y$ is the class label. During the learning process, its objective is to minimize the spherical distance of feature representations with predefined basis vectors within such a set $\mathcal{W}$.

\section{Methodology}
\begin{figure}[tbp]
	\centering
	\includegraphics[width=1.00\columnwidth]{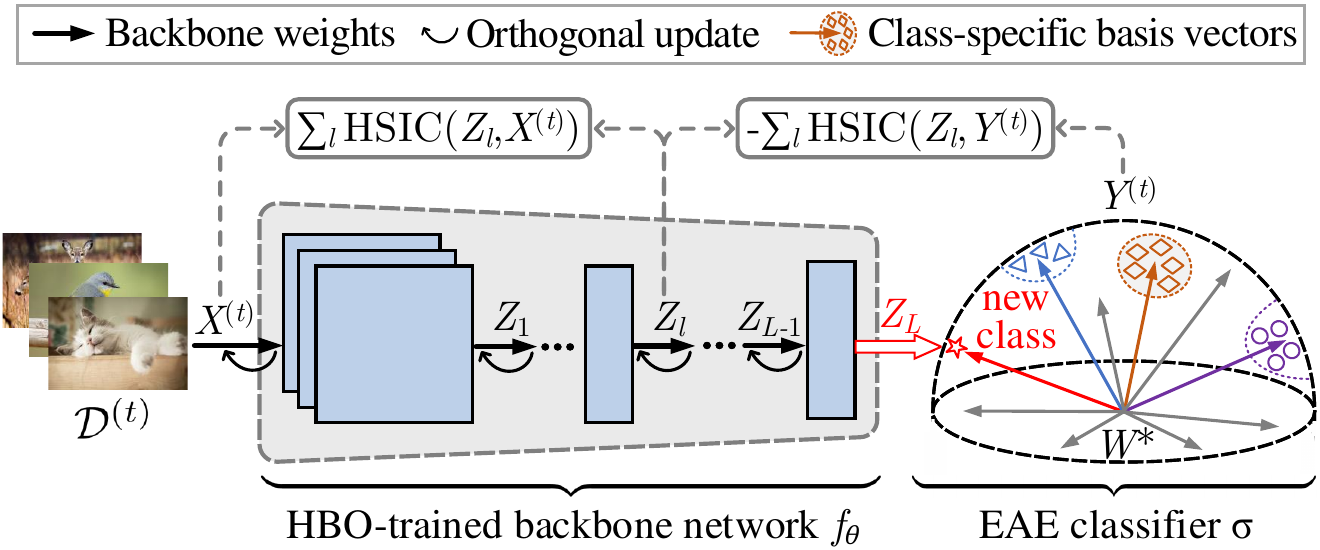} 
	\caption{Overview of CLDNet. HBO transforms learning task $t$ into a constrained statistical dependency mini-max problem and EAE predicts by matching class-specific basis vectors. Systematically, the last-layer hidden representation $Z_L$ is bound to any one of the available basis vectors in grey for recognizing a new class. We mark this process in red.}
	\label{Fig_Overview}
\end{figure}

We propose a conceptually simple yet effective method that reaches multiple Continual Learning Desiderata within a single Network (CLDNet). This is accomplished by the synergy between two key components: HBO implements non-overwritten parameter updates mediated by the Hilbert-Schmidt independence criterion in an orthogonal space, and EAE enhances decision boundary adaptation between old and new tasks with predefined basis vectors. As depicted in Figure~\ref{Fig_Overview}, our model is formulated by two nested parts: the backbone network $f_\theta$ and the parameter-free classifier $\sigma$; For a single input $x$, we have its output $o(x) = \sigma(f_\theta(x))$. The following explains how CLDNet learns continually via the statistical dependency and distance metric.

Before elaboration, some definitions related to CL setting are introduced as follows. CL trains a model incrementally on a sequence of task datasets $\{\mathcal{D}^{(1)}, \mathcal{D}^{(2)}, \dots, \mathcal{D}^{(T)}\}$, where $\mathcal{D}^{(t)} = \{(x_i^{(t)},y_i^{(t)})_{i=1}^{|\mathcal{D}^{(t)}|}\}$ (or denoted by $X^{(t)}$ and $Y^{(t)}$) and $|\mathcal{D}^{(t)}|$ is the number of samples from task $t$. The output class $\mathcal{C}^{(t)}$ has no overlap, i.e., $\mathcal{C}^{(t)}\cap \mathcal{C}^{(t')} = \varnothing (t\neq t')$. Once a task is learned, its training data is often no longer accessible. At inference time, we mainly focus on one of the most challenging class-IL scenarios where task identities of test instances from classes seen so far are unknown.

\subsection{HSIC-Bottleneck Orthogonalization (HBO)}
\textit{We first present the learning objective of the HBO process. We then iteratively calculate the orthogonal projector facilitated by HSIC terms in batch learning form.}

Suppose we have the backbone network $f_\theta$ with $L$ hidden layers activated by the function $S_l(\cdot): \mathbb R^{d_{l-1}}\rightarrow \mathbb R^{d_l}$, yielding hidden representations $Z_l\in\mathbb R^{m\times d_l}$ ($l=1,2,\dots,L$), where $m$ is the batch size. Denote by $\theta_l$ the parameter to be updated, $Z_l = S_l(Z_{l-1}\theta_l)$. Following OWM~\cite{NMI2019OWM}, we adopt the orthogonal projector\footnote{This form is equivalent to $P_l=I-A_l(A_l^\mathrm{T} A_l+\alpha I)^{-1}A_l$ used in OWM, where $A_l$ consists of all learned hidden representations $Z_{l-1}$ and $\alpha$ is a small constant.} $P_l=\alpha(A_l^\mathrm{T} A_l+\alpha I)^{-1}$ that regularizes gradient update directions orthogonal to  the input of previous tasks. The difference lies in that we recast it via the HSIC-bottleneck training. Then, HBO transforms CL into a constrained statistical dependency problem:
\begin{equation}\label{HBO}
	\begin{split}
		\min_{Z_l}\!&: \sum_{l=1}^L{\rm HSIC}(Z_l,X^{(t)}) - \beta{\rm HSIC}(Z_l, Y^{(t)})\\
		{\rm s.t.}\!&: \theta_l^t = \theta_l^{(t-1)} - \lambda P_l^{\rm HSIC}\Delta\theta_l^{(t-1)}, P_l^{\rm HSIC}A_l=O 
	\end{split}
\end{equation}

\noindent where $\beta$ is the balancing factor of HSIC terms, $\Delta\theta_l^{(t-1)}$ is the parameter update via HSIC-bottleneck training, $\lambda$ is the learning rate, and $ P_l^{\rm HSIC}$ is the orthogonal projector for modulating gradients. Unlike OWM, we construct it by capturing nonlinear dependencies of different layers. As  formulated by Equation (\ref{HBO}), for each layer of backbone network $f_\theta$, the dependence between the input $X^{(t)}$ and hidden representations $Z_l$ is minimized while that between the output $Y^{(t)}$ and hidden representations $Z_l$ is maximized. 

To gain insights into HBO, we want to emphasize that the optimal hidden representation $Z_l$ is more amenable to CL, which inherently contributes to non-overwritten parameter updates in the perspective of reducing \textit{feature bias}: 

The conventional cross-entropy loss learns more on discriminative features (e.g., the dominant parts) that can recognize the classes of the task~\cite{AAAI2021PCL}. During sequential training, some of those not previously learned features (e.g., the non-dominant parts) may become dominant for recognizing the new task classes, resulting in feature bias in the backbone network, as revealed by the recent study~\cite{guo2022online}. By contrast, HBO encourages learning all possible features from a sequence of tasks, implying that some of the features that may not be sufficiently discriminative for the current task are also holistically considered. This is achieved by optimizing hidden representation $Z_l$ layer by layer to seek a balance between independence from unnecessary details of the input $X^{(t)}$ and dependence on the output $Y^{(t)}$. In this sense, the information needed to predict the label is well acquired and permeated in $Z_l (l=1,2,\dots, L)$ when Eq. (\ref{HBO}) converges. This not only mitigates the feature bias but also constructs a more accurate projector, thereby facilitating non-overwritten parameter updates. With this pitfall addressed, we achieve significant accuracy gains over OWM in empirical evaluation.

Now let us compute the orthogonal projector using the mini-batches of data, denoted by $Z_0(k) = X^{(t)}(k)$, $Z_l(k) = [z_l^1,z_l^2,\dots,z_l^k]$. We embed the recursive least square algorithm~\cite{golub2013matrix} into HSIC-bottleneck training. The derivation progress is based on Woodbury matrix identity, from which we get the following iterative expression:
\begin{equation}\label{P_kk}
	\begin{split}    
		P_l^{\rm HSIC}(k+1) =& P_l^{\rm HSIC}(k)- \\ &\dfrac{P_l^{\rm HSIC}(k) z_l^{k+1}(z_l^{k+1})^\mathrm{T}P_l^{\rm HSIC}(k)} {\alpha + (z_l^{k+1})^\mathrm{T}P_l^{\rm HSIC}(k)z_l^{k+1}}
	\end{split}
\end{equation}

We make two remarks about the above derivation. (1) Equation (\ref{P_kk}) circumvents the matrix-inverse operation in the original matrix form. Meanwhile, both inputs and learned hidden representations from previous tasks are no longer revisited; Instead only the most recently updated $P_l^{\rm HSIC}(k)$ is required. (2) As mentioned in related work, AOP improves OWM itself by converting the small constant $\alpha$ into an adaptive one associated with each training batch. However, our work differs significantly in terms of motivation and methodology, and optionally, incorporates it for hyper-parameter selection, as is demonstrated in the experiments. 

\subsection{EquiAngular Embedding (EAE)}
\textit{This part first describes the indispensability of the EAE process for decision boundary adaptation and then elaborates on how it works together with HBO.} 

In our CLDNet, the HBO-trained backbone network outputs its last-layer hidden representation, which contains the information necessary for decision, but not necessarily in the available form like the logit of each class. One simple way is to append a single output layer (or projection head) trained with softmax cross-entropy~\cite{ma2020hsic}. However, this fully-connected layer would be added in a fully parametric manner, requiring additional consideration of parameter overwriting issues; Importantly, the decision boundary/output space of old tasks would be squeezed by the new task. 
To tackle these, inspired by the recently proposed EBVs~\cite{shen2023equiangular}, we design a novel EquiAngular Embedding (EAE) which replaces the trainable classifier parameters with predefined basis vectors. 

EAE starts with a data-independent predefined process. Recall that, in the preliminary part, we denote by $\mathcal{W} = \{w_c\}_{c=1}^C$ the set composed of $C$ $d$-dimensional basis vectors on the surface of a unit hypersphere $S^d \in \mathbb{R}^d$, and $\gamma \in [0,1)$ the common angle pairwise separating these basis vectors. The question arises of how to construct such a set $\mathcal{W}$ which satisfies Equation (\ref{EBV_Def}) when given fixed $\gamma$, $d$, and $C$. Specifically, we randomly
initialize a matrix $W \in\mathbb R^{d\times C}$ with normalized rows such
that the angle between any two vectors $\arccos(w_i, w_j)$ equals $w_i^\mathrm{T}\cdot w_j$, in which $w_i, w_j\in\mathbb R^d$ ($w_i\leftarrow\frac{w_i}{|w_i|}$, $i, j = 1,2,\dots, C$, $i\neq j$). Then, we further tweak unit basis vectors in  $W$ by the following optimization function~\cite{shen2023equiangular}:  
\begin{equation}\label{EAE_generation}
	W^* = \arg\min_{W}\sum\nolimits_{i=1}^{N-1}\sum\nolimits_{j > i}^N \max(|w_i^\mathrm{T}\cdot w_j|-\gamma, 0)
\end{equation}

\noindent This formulation cuts out the gradient of those unit vector pairs that hold $-\gamma \leq w_i^\mathrm{T}\cdot w_j \leq \gamma$ and optimizes the remaining ones. 

Now let us consider EAE in the context of CL. When it comes to decision-making, classes from different tasks would be sequentially bound to some $w_c^*\in\mathbb R^d$ in $W^*\in\mathbb R^{d\times C}$, allowing models to scale to a large number of possible outputs, without a linear cost in the number of parameters.
For a single input $x_i^{(t)}$ of task $t$, we have the backbone network output $z_i^{(t)}$ of the $L$th-layer (we omit the subscript $L$ for brevity); According to Equation (\ref{EBV_Dis}), the probability of $z_i^{(t)}$ recognized as the class $y_i^{(t)}$ can be rewritten as:
\begin{equation}\label{EAE_Prob}
	P(y=y_i^{(t)}|z_i^{(t)}) = \frac{{\rm exp}(z_i^{(t)}\tilde{w}_c^*)}{\sum_{c'=1}^{\mathcal{C}^{(t)}}{\rm exp}(z_i^{(t)}\tilde{w}_{c'}^*)}
\end{equation}

\noindent where $\tilde{w}_c^*$ denotes the $\ell_2$-normalized $w_c^*$. Intuitively, this is equivalent to optimizing the cosine similarity between each $z_i^{(t)}$ and $\tilde{w}_c^*$. Therefore, based on Equations (\ref{HBO}) and (\ref{EAE_Prob}), the overall training objective of our CLDNet can be converted into minimizing the negative log-likelihood over task $t$:
\begin{equation}\label{CLDNet_Obj}
	\min_\theta: -\frac{1}{|\mathcal{D}^{(t)}|}\sum_{(x_i^{(t)},y_i^{(t)})}\log P(y=y_i^{(t)}|f_{\theta}(x_i^{(t)})) 
\end{equation}

\noindent where $|\mathcal{D}^{(t)}|$ is the number of training samples from task $t$. Therefore, as an alternative to logits, the prediction is made by keeping the input $x_i^{(t)}$ with representation $z_i^{(t)}$ as close as possible to its class-specific basis vectors. Owe to our unique training objective, we only use a simple loss without storing any old task exemplars or backbones, suggesting that our CLDNet differs from most existing CL methods significantly. Please refer to Algorithm~\ref{alg:algorithm_1} 
for more.


\begin{algorithm}[t]
	\caption{CLDNet Training and Test algorithm}
	\label{alg:algorithm_1}
	\textbf{Input}: A sequence of task datasets $\{\mathcal{D}^{(1)}, \mathcal{D}^{(2)}, \dots, \mathcal{D}^{(T)}\}$, backbone network $f_\theta$ with $L$ hidden layers, followed by the parameter-free classifier $\sigma$, learning rate $\lambda$, batch size $m$, etc.\\ 
	\textbf{Output}: $\sigma(f_\theta(\cdot))$
	\begin{algorithmic}[1] 
		\STATE \textbf{\textit{\# predefine basis-vector matrix prior to training}} \\
		\STATE Initialize $W\in\mathbb R^{d\times C}$ with normalized rows; \\
		\STATE Obtain $W^*$ by tweaking $W$ with  Equation~(\ref{EAE_generation}); \\
		\STATE \textbf{\textit{\# during sequential training in batch learning}} \\
		\FOR{$t=1, 2,\dots,T$}
		\FOR{$j=1, 2,\dots,|\mathcal{D}^{(t)}|/m$}
		\STATE Calculate hidden representations $\{Z_l\}_{l=1}^L$;
		\STATE Assign last-layer hidden representation $Z_L$ class-specific basis vectors from $W^*$; 
		\STATE Solve the constrained statistical dependency problem with Equation~(\ref{HBO});
		\STATE Update the backbone network parameter $\theta$ with Equation~(\ref{CLDNet_Obj});
		\ENDFOR
		\ENDFOR
		\STATE \textbf{\textit{\# at test time}} \\
		\STATE Draw test instances from any of tasks 1 to $T$;
		\STATE \textbf{return} predicted labels by retrieving their closest class-specific basis vectors.
	\end{algorithmic}
\end{algorithm}

\subsection{Reaching Continual Learning Desiderata}
Here we discuss how CLDNet reaches multiple CL desiderata. The above formulations in our method clearly require no \textit{rehearsal buffer} and substantial \textit{network expansion}, which are non-trivial. For example, although a limited rehearsal buffer is allowed in the CL community, some prior works opt for sufficiently large buffer
sizes that even suffice to train a supervised counterpart, as revealed by GDumb~\cite{prabhu2020gdumb}. The same goes for arbitrarily expanding task-specific backbone networks, which would result in misleading high-accuracy performance. We think understanding this question is very important for future research, e.g., one should count these \textit{non-desiderata} into the memory budget for a fair comparison~\cite{zhou2022model}.

Focusing on such a strict yet realistic setting, it is natural to think of whether our CLDNet achieves the stability and plasticity trade-off. We would still like to emphasize that this balance is realized by the synergy of HBO and EAE. On the one hand, HBO addresses the layer-wise parameter overwriting in the backbone network, followed by the parameter-free EAE classifier. \textit{This ensures the persistent knowledge retention of past tasks.} On the other hand, the rank of orthogonal projectors theoretically matters in the backbone network capacity available for incoming tasks and we construct such a nonzero matrix $P_l^{\rm HSIC}(k)$, allowing for some degree of freedom to learn new tasks; Meanwhile, since predefined basis vectors are exactly equivalent, the output space is no longer constrained to the number of classes. \textit{This maintains the required network capacity to accommodate new tasks.}

\section{Empirical Evaluation}
We perform extensive experiments to evaluate the proposed CLDNet in the challenging class-IL setting. First, we introduce the experimental setup. We then provide the experimental results and discussion, following which we conduct ablation studies on the core components in our algorithm.

\subsection{Experiment Setting}
\textbf{Dataset and Split.}
We experiment on multiple evaluation benchmarks for class-IL. \textbf{Small Scale:} MNIST~\cite{lecun1998gradient} contains 60,000 handwritten digit images in the training set and 10,000 samples in the test set, which is split into 5 disjoint tasks with 2 classes per task; FashionMNIST~\cite{FashionMNIST} is an MNIST-like fashion product benchmark where the ten objects are split into five two-class classification tasks; CIFAR-10~\cite{CIFAR-100} has 10 classes with 50,000 samples for training and 10,000 for testing, which is divided into 5 tasks with 2 classes per task. 
\textbf{Medium Scale:} CIFAR-100~\cite{CIFAR-100} comprises 60,000 images belonging to 100 distinct classes, which are further divided into 10 tasks with each task containing 10 disjoint classes. 
\textbf{Large Scale:} ImageNet-R~\cite{hendrycks2021many} has 200 classes with 24,000 samples for training and 6,000 for testing. It is split into 10 tasks with 20 classes in each task. ImageNet-R incorporates newly curated data encompassing diverse styles, such as cartoons, graffiti, and origami, alongside challenging examples from ImageNet that conventional models (e.g., ResNet) fail to recognize. The substantial intra-class variability renders it more akin to intricate real-world problems.

\textbf{Training Details.}
\textbf{Architectures:} In our experiments, all methods use similar-sized neural network architectures. For MNIST and FashionMNIST, following the setting in \cite{wolczyk2022continual}, we use a standard MLP with 2 hidden layers of size 400; 
For CIFAR-10, following the setting in \cite{NMI2019OWM, guo2022adaptive}, we use a CNN with 3 convolutional layers; For CIFAR-100, following the similar setting in \cite{bonicelli2022effectiveness, wang2023dualhsic}, we use a wide-adopted ResNet18; 
For ImageNet-R, following the setting in \cite{wang2022dualprompt}, we use the ViT-B/16 pre-trained on ImageNet and allow all methods to start from the same pre-training for a fair comparison. \textbf{Hyper-parameters:} We either reproduce results using suggested hyper-parameters in their source code repositories or directly take existing results reported in state-of-the-art (SOTA) baselines. In our CLDNet, for HBO we set the coefficient $\beta = 500$ and adopt the Gaussian kernel as suggested by \cite{ma2020hsic}, as well as the adaptive $\alpha$ with an initial value 0.01 for the orthogonal projector, like~\cite{guo2022adaptive}; For EAE we set $\gamma=0.04$, $d=1000$, and $C=1000$ following the recommendations by EBVs~\cite{shen2023equiangular}.  \textbf{Computing Infrastructure:} All experiments are run in PyTorch using NVIDIA RTX 3080-Ti GPUs with 12GB memory.

\subsection{Results and Comparison}
This paper considers a strict yet realistic setting for defying forgetting, covering three aspects of CL desiderata: ideally (i) accessing no training data of previous tasks, (ii) maintaining the model size relatively unchanged during sequential training, and (iii) striking a balance between stability and plasticity. To demonstrate the superiority of our work, we extensively compare it with the representative and SOTA competitors. Since different methods have very different requirements in data, networks, and computation, it is intractable to compare all in the completely same experimental conditions. Therefore, we compare our CLDNet with regularization-, rehearsal-, and architecture-based approaches respectively.

\textbf{Comparison with Regularization-based Approaches.} 
Table~\ref{table:regularization} compares our CLDNet with regularization-based approaches, which typically penalize parameter variations over an over-parameterized network by regularizers or orthogonal projectors. Without rehearsal buffers or network expansion, class-IL is particularly difficult for these methods. The competitors include EWC~\cite{PNAS2017EWC}, SI~\cite{ICML2017SI}, MAS~\cite{ECCV2018MAS}, OEWC \cite{ICML2018Online_EWC}, OWM~\cite{NMI2019OWM}, DMC~\cite{WACV2020DMC}, ICNet~\cite{wolczyk2022continual}, and more recent SOTA methods, AOP~\cite{guo2022adaptive}, and CRNet~\cite{li2023CRNet}. During class-IL, methods like EWC, MAS, OEWC, and SI struggle with the stability and plasticity dilemma as it is difficult to correctly assign credit to weights with the number of tasks increasing. Similar findings can be observed in~\cite{wolczyk2022continual}. Recent AOP and CRNet are strong baselines, but our method consistently outperforms them on all three task sequences by clear margins. Compared with the second-best results, CLDNet gets an improvement of 2.16\% 4.39\%, and 2.56\% on the Split MNIST, FMNIST, and CIFAR-10, respectively.  

\begin{table}[t]
	\centering
	\resizebox{0.93\columnwidth}{!}{
		\begin{tabular}{lccc}
			\toprule
			Method & MNIST &  FMNIST & CIFAR-10 \\
			\midrule
			EWC & 36.52 $\pm$ 2.54 & 35.16 $\pm$ 5.33 & 18.92 $\pm$ 4.88       \\ 
			MAS  & 38.74 $\pm$ 2.67 & 33.78 $\pm$ 6.42      & 17.79 $\pm$ 6.04  \\ 
			OEWC & 40.52 $\pm$ 6.84  & 38.17 $\pm$ 4.02     & 16.98 $\pm$ 5.21     \\
			SI  &45.28 $\pm$ 0.57 & 40.23 $\pm$ 3.34         & 17.38 $\pm$ 4.13     \\
			ICNet & 40.73 $\pm$ 3.26 & 35.11 $\pm$ 0.02 & 19.07 $\pm$ 0.15  \\ 
			DMC & 90.76 $\pm$ 0.25 & 72.54 $\pm$ 1.25 & 51.28 $\pm$ 0.95 \\
			OWM & 91.60 $\pm$ 0.13 & 80.32 $\pm$ 0.73 & 52.83 $\pm$ 0.87 \\
			AOP & 94.43 $\pm$ 0.21 & 82.97 $\pm$ 0.95 & 53.56 $\pm$ 0.29 \\
			CRNet & 94.45 $\pm$ 0.36 & 90.98 $\pm$ 0.83 & 50.01 $\pm$ 0.58 \\ \midrule
			Ours &\bf 96.61 $\pm$ 0.15 &\bf 95.37 $\pm$ 0.68 &\bf 56.12 $\pm$ 0.33 \\
			\bottomrule
		\end{tabular}
	}
	\caption{Average accuracy (\%) across all five tasks of the Split MNIST, FashionMNIST (FMNIST), and CIFAR-10, evaluated after learning the whole sequence. All methods are run 5 times, with the mean and standard deviation reported.}
	\label{table:regularization}
\end{table}

\begin{table}[t]
	\centering
	\resizebox{0.96\columnwidth}{!}{
		\begin{tabular}{lccc}
			\toprule
			& \multicolumn{3}{c}{Buffer size for CIFAR-100} \\ \cmidrule{2-4}
			Method & 200 & 500 & 2000 \\ \midrule
			ER &18.09$\pm$1.33  &28.25$\pm$0.69  &43.18$\pm$2.00  \\
			X-DER-RPC &51.40$\pm$2.17  &57.45  &62.46  \\
			ER-ACE &41.85$\pm$0.83  &48.19  &57.34  \\
			DER++ &26.25$\pm$0.96  &43.65  &58.05  \\
			OCM &52.08$\pm$1.13  &56.93$\pm$0.86  &61.79$\pm$0.42  \\
			LiDER &51.23$\pm$1.93  &57.76$\pm$0.75  &62.78$\pm$0.51  \\
			DualHSIC &52.67$\pm$1.81  &57.88$\pm$1.04  &62.70$\pm$0.57  \\ \midrule
			Ours & \multicolumn{3}{c}{\textbf{65.42$\pm$0.36}  (Buffer size = 0)} \\
			\bottomrule
		\end{tabular}
	}
	\caption{Average accuracy (\%) across all ten tasks of the Split CIFAR-100. All results except ours, OCM, and LiDER are from \cite{wang2023dualhsic}. LiDER and DualHSIC are built upon X-DER-RPC to report their best performance.}
	\label{table:rehearsal}
\end{table}

\textbf{Comparison with Rehearsal-based Approaches.} 
As reported in Table~\ref{table:rehearsal}, we evaluate our CLDNet with rehearsal-based approaches on CIFAR-100, which assume access to partial old data. The competitors include ER~\cite{chaudhry2019tiny}, DER++~\cite{buzzega2020dark}, X-DER-RPC~\cite{boschini2022class}, ER-ACE~\cite{caccia2021new}, more recent SOTA methods, OCM~\cite{guo2022online}, LiDER~\cite{bonicelli2022effectiveness}, and DualHSIC~\cite{wang2023dualhsic}. Both LiDER and DualHSIC serve to improve the performance of rehearsal-based counterparts. We observe that these baselines deteriorate when buffer size decreases, which may paralyze when buffer size is zero. By contrast, the performance gains of CLDNet are 12.75\%, 7.54\%, and 2.64\% from buffer size 200 to 2000, respectively, thanks to the synergy between HBO and EAE in our CLDNet.

Table~\ref{table:rehearsal_pre} further compares with rehearsal-based approaches on challenging ImageNet-R, including BiC~\cite{wu2019large}, GDumb~\cite{prabhu2020gdumb}, Co$^2$L~\cite{cha2021co2l}. Since we following the pre-training used in L2P~\cite{wang2022learning} and DualPrompt~\cite{wang2022dualprompt}, these two prompt-based methods are also considered. This corresponds to two versions of the proposed method: Ours(1) refers to only one pre-trained ViT being used while Ours(2) involves two ViT like L2P and DualPrompt. We observe that the performance of rehearsal-based methods exhibits an obvious decline as the buffer size decreases---the significant intra-class diversity of ImageNet-R poses a great challenge for rehearsal-based methods to work effectively with the buffer size of 1000. This suggests again the necessity of CLDNet as a rehearsal-free method. Compared with prompt-based methods, Ours(1) beats L2P by 3.80\% and falls short of DualPrompt by 2.76\%; Ours(2) surpasses DualPrompt by 3.30\%. This additionally indicates that our method can accommodate the real-world scenario where pre-training is usually involved as a base session.


\begin{table}[t]
	\centering
	\resizebox{0.95\columnwidth}{!}{
		\begin{tabular}{lccc}
			\toprule
			& \multicolumn{3}{c}{Buffer size for ImageNet-R} \\  \cmidrule{2-4}
			Method & 0 & 1000 & 5000 \\ \midrule
			ER &-&55.13$\pm$1.29  &65.18$\pm$0.40  \\
			BiC &- &52.14$\pm$1.08 &64.63$\pm$1.27  \\
			GDumb &- &38.32$\pm$0.55  &65.90$\pm$0.28  \\
			DER++ &- &55.47$\pm$1.31  &66.73$\pm$0.81  \\
			Co$^2$L&-  &53.45$\pm$1.55  &65.90$\pm$0.14  \\
			L2P &61.57$\pm$0.66  & - & - \\
			DualPrompt &68.13$\pm$0.49  &- &-  \\
			Ours(1) &65.37$\pm$0.39  &- &-  \\
			Ours(2) &\bf 71.43$\pm$0.22  &- &-  \\ \midrule
			Upper bound & &79.13$\pm$0.18  &  \\
			\bottomrule
		\end{tabular}
	}
	\caption{Average accuracy (\%) across all ten tasks of the Split ImageNet-R. When buffer size = 0, “-” denotes most rehearsal-based methods are not applicable anymore; When buffer size = 1000 or 5000, “-”  denotes the omitted results.}
	\label{table:rehearsal_pre}
\end{table}

\textbf{Comparison with Architecture-based Approaches.}
To make the comparison more complete, we also compare our CLDNet with architecture-based (i.e., network expansion) approaches, which assign new branches for each task. The competitors include PNN~\cite{rusu2016progressive}, DEN~\cite{yoon2018lifelong}, RCL~\cite{xu2018reinforced}, APD~\cite{yoon2020scalable}, PCL~\cite{AAAI2021PCL}. Table~\ref{table:architecture} reports the results on CIFAR-100. CLDNet achieves the best Avg. Acc 64.99\% with minimal Capacity 1.02$\times$. For a fair comparison, we do not compare with hybrid methods that deviate significantly from the CL desiderata that our method is designed for: this excludes methods that heavily rely on both network expansion and rehearsal buffers. On the one hand, hybrid methods like RPS-Net and EDR do push the performance towards the upper bound achieved by offline training. On the other hand, e.g., for CIFAR-100, they explicitly use a buffer size of 2000 and require about 5$\times$ (RPS-Net) or 10$\times$ (DER w/o P) more parameters than the base network.  

\begin{table}[t!]
	\centering
	\resizebox{0.80\columnwidth}{!}{
		\begin{tabular}{lccc}
			\toprule
			Method & Buffer size & Capacity & Avg. Acc \\ \midrule
			PNN & \multirow{6}{*}{0} &1.71$\times$  &54.90$\pm$0.92 \\
			DEN &  &1.81$\times$ &57.38$\pm$0.56 \\
			RCL &  &1.80$\times$ &55.26$\pm$0.13 \\
			APD &  &1.53$\times$ &61.18$\pm$0.20 \\
			PCL &  &1.46$\times$ &62.58$\pm$0.32 \\
			Ours &  &\bf 1.02$\times$ &\bf 64.99$\pm$0.24 \\
			\bottomrule
		\end{tabular}
	}
	\caption{Performance comparison on the Split CIFAR-100. The metric Capacity (lower is better) measures what extent a model scales after learning the whole sequence using the convolutional architecture in~\cite{yoon2020scalable}.}
	\label{table:architecture}
\end{table}

\begin{figure}[t!]
	\centering
	\includegraphics[width=0.70\columnwidth]{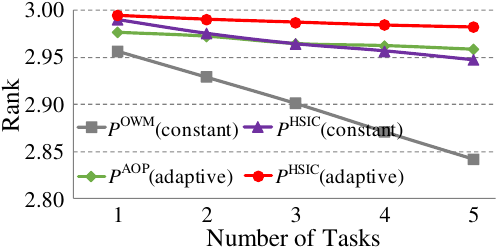} 
	\caption{Changes of the rank of orthogonal projectors.}
	\label{Fig_Rank}
\end{figure}

\begin{figure}[t]
	\centering
	\includegraphics[width=0.95\columnwidth]{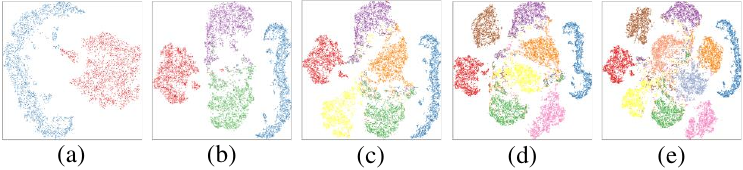}
	\caption{t-SNE visualization based on split FashionMNIST. Each color represents a class. We visualize two classes in each task as a session. (a)-(e) represents the corresponding representation visualization of classes trained so far.}
	\label{Fig_tSNE}
\end{figure}

\textbf{Ablation Study.}
Figure~\ref{Fig_Rank} plots the rank of different orthogonal projectors under a \textit{constant} or \textit{adaptive} $\alpha$, where the same MLP with 3 hidden layers is trained for the split MNIST. Interestingly, AOP improves OWM by shrinking the change in rank as tasks increase; The HSIC used in our method contributes to a higher rank for maintaining model plasticity. In addition, we provide an empirical analysis of the effectiveness of HBO for learning (see Figure~\ref{Fig_tSNE}) and EAE for decision (see Table~\ref{table:ablation}). In summary, both components contribute to the final performance improvement, e.g., parameter-free EAE classifier $\sigma$ facilitates decision boundary adaptation.

\begin{table}[t!]
	\centering
	\resizebox{0.95\columnwidth}{!}{
		\begin{tabular}{lccc}
			\toprule
			Component & MNIST & FMNIST & CIFAR-10 \\ \midrule
			HBO + $\sigma_{\theta'}$ & 19.96 & 19.74 & 15.56 \\
			HBO + $\sigma_{\theta''}$  & 95.12 & 92.77 & 53.89 \\
			HBO +  $\sigma$ (i.e., EAE) &\bf 96.62 &\bf 95.35 &\bf 56.10 \\
			\bottomrule
		\end{tabular}
	}
	\caption{Effectiveness of the core designs in our CLDNet. $\sigma_{\theta'}$ ($\sigma_{\theta''}$) represents a single fully-connected output layer, parameterized by $\theta'$ ($\theta''$), and is trained by cross-entropy loss without (with) an orthogonal projector for prediction.}
	\label{table:ablation}
\end{table}


\section{Conclusion}
This present study considers a stringent yet practical setting to reach multiple CL desiderata. Taking the statistical dependency and distance metric as training objectives, we propose CLDNet with two pivotal components, i.e., HBO for non-overwritten parameter updates and EAE for decision boundary adaptation. We perform extensive experiments to show that CLDNet achieves a better stability-plasticity trade-off in a rehearsal-free and minimal-expansion way. Moreover, we hope that our study inspires further research in reaching CL desiderata, e.g., this makes sense for real-world applications under privacy-sensitive and resource-limited CL scenarios.

\section*{Acknowledgments}
This work was supported by the National Key R\&D Program of China under Grant 2021ZD0201300, the National Natural Science Foundation of China under Grants U1913602 and 61936004, and the 111 Project on Computational Intelligence and Intelligent Control under Grant B18024.

\bibliography{aaai24}

\end{document}